\pdfoutput=1
\documentclass{article}
\usepackage[english]{babel}
\usepackage[utf8]{inputenc}
\usepackage[letterpaper,top=2cm,bottom=2cm,left=3cm,right=3cm,marginparwidth=1.75cm]{geometry}
\usepackage{stmaryrd}
\usepackage{color,amsmath,amssymb, amsfonts, amstext,amsthm, latexsym}
\usepackage{booktabs}
\usepackage{multirow}
\usepackage{amsmath}
\usepackage{amsthm}
\usepackage{flushend, cuted}
\usepackage[title]{appendix}
\usepackage{bbm}
\usepackage{float}
\usepackage{epsfig, graphicx, graphics}
\usepackage{longtable}
\usepackage{cite}
\usepackage{subfigure}
\usepackage{caption}
\usepackage{bm}
\usepackage{authblk}
\usepackage[mathscr]{euscript}
\usepackage{diagbox}
\usepackage[ruled,vlined]{algorithm2e}
\usepackage{multirow}

\newtheorem{definition}{Definition}

\newtheorem{example}{Example}

\title{Learning effective dynamics from data-driven stochastic systems}

\author[a,b]{Lingyu Feng 
}
\author[a,b]{Ting Gao \thanks{Corresponding author: tgao0716@hust.edu.cn }}
\author[c]{Min Dai 
}
\author[d]{Jinqiao Duan 
}

\affil[a]{School of Mathematics and Statistics, Huazhong University of Science and Technology, Wuhan 430074, China}
\affil[b]{Center for Mathematical Science, Huazhong University of Science and Technology, Wuhan 430074, China}
\affil[c]{School of Science, Wuhan University of Technology, Wuhan 430070, China}
\affil[d]{College of Science, Great Bay University, Dongguan, Guangdong 523000, China}

\begin{document}

\maketitle

\begin{abstract}
Multiscale stochastic dynamical systems have been widely adopted to a variety of scientific and engineering problems due to their capability of depicting complex phenomena in many real world applications. This work is devoted to investigating the effective dynamics for slow-fast stochastic dynamical systems. Given observation data on a short-term period satisfying some unknown slow-fast stochastic systems, we propose a novel algorithm including a neural network called Auto-SDE to learn invariant slow manifold. Our approach captures the evolutionary nature of a series of time-dependent autoencoder neural networks with the loss constructed from a discretized stochastic differential equation. Our algorithm is also validated to be accurate, stable and effective through numerical experiments under various evaluation metrics. 

\end{abstract}


\begin{quotation}
Multiscale modeling is ubiquitous in many real world applications and is often studied with various methods. Though modeling such complex systems is challenging, it is still attractive to many researchers, due to the importance of studying its intrinsic mechanisms. Things become even harder when the stochastic system is unknown or partially known, and the observation data is limited within a short time period. In this case, to investigate the effective structure of an unknown slow-fast stochastic system, we introduce a reliable algorithm with time evolutionary auto-encoder neural network. Combined with stochastic numerical integrator, we compute the approximated manifold through a short-term period dataset and then the slow-fast system can be reduced to a slow system.
\end{quotation}

\section{Introduction}
\noindent 

Numerous complex systems in the areas of science, engineering, chemistry or material science have the philosophy of multiscale properties in their dynamic evolution \cite{Fowler1996TheLS,Goussis2013TheRO,Carpenter2018EmpiricalCO,Zhang2021ChemicalKM}. By considering models at different scales simultaneously, we would like to obtain both the efficiency of the macroscopic models as well as the accuracy of the microscopic models. For example, approaches in chemistry usually involve the quantum mechanics models in the reaction region and the classical molecular models elsewhere \cite{Weinan2011PrinciplesOM}. Besides, as noisy observations always exist in all kinds of systems under internal or external factors, stochastic dynamical systems come to play an important role in modeling such phenomena. Thus, it is of great importance to study multiscale stochastic dynamical systems \cite{Weinan2011PrinciplesOM,Kuehn2015MultipleTS}. To better understand the intrinsic nature of such complex systems, researchers usually try to investigate the effective dynamics of these systems, such as invariant manifolds, global attractors, tipping points, noise induced bifurcations, transition pathways, and so on\cite{stratonovich1967topics,Arnold2003RandomDS,Berglund2005NoiseInducedPI,pavliotis2008multiscale, Duan2015AnIT}. These dynamical behaviors could capture the fundamental structures when the system evolves over time or parameter space. However, there are still challenges in simulating and analyzing such systems with the conventional methods due to the curse of high dimensionality \cite{donoho2000high}. Thus, in this paper, we will formulate a general deep learning framework to learn effective dynamics evolving on the invariant manifold given short-term observation data and an unknown system.

There are various traditional ways to handle invariant manifolds for deterministic dynamical systems. To find parameterization of the invariant manifold, with the knowledge of geometry, different approaches appear to compute global stable or unstable manifold of a vector field\cite{Krauskopf2005ASO}. Specifically, the approaches include geodesic level sets, boundary value problem of trajectories, fat trajectories, partial differential equation formulation, and box covering algorithm. From the computational side, an adaptive method and a finite-element method are also presented to compute invariant manifold\cite{Branicki2009AnAM,Jain2021HowTC} and a kernel method is offered for center manifold approximation\cite{haasdonk2021kernel}. Restricting to the invariant manifold, multiscale systems could reduce to the effective systems\cite{Gear2005ProjectingTA}. For instance, Gear et al. use singular perturbation expansion with respect to the scaling factor to approximate the invariant slow manifold for any dimension. 

For slow-fast stochastic dynamical systems, the invariant manifold also contributes to obtaining effective systems. On one hand, the stochastic system can be converted to the random system which admits a random invariant manifold by the Lyapunov-Perron integral equation and then the lower dimensional system follows\cite{Duan2003INVARIANTMF,Schmalfu2008InvariantMF,Ren2012ApproximationOR,Fu2012SlowMF}. Another popular way to acquire a reduced system, without computing invariant manifold, is averaged equation in terms of invariant measure and the averaging principle \cite{khasminskij1968principle,namachchivaya2000stochastic,Wang2009SlowMA,liu2020averaging}. However, there is some limitation to this method when the invariant measure for each system is hard to obtain. In this circumstance, data mining and nonlinear stochastic model reduction techniques are adopted to learn a low dimensional invariant manifold \cite{Dsilva2016DataDrivenRF,Ye2021NonlinearMR}. For example, in \cite{Dsilva2016DataDrivenRF}, the authors utilize both the local geometry and the local noisy dynamics through diffusion maps to detect the slow components with multiscale data. 

With the development of modern computation techniques, deep learning is becoming more and more popular, which benefits a variety of application fields. The interdisciplinary study of dynamical systems and deep learning has great advantages. On the one hand, deep learning methods encourage effectively calculating a lot of complex dynamical behaviors from data, especially for high-dimensional case\cite{Vlachas2022MultiscaleSO,Krajvnak2021PredictingTB,manoni2020effective,lu2017data}. On the other hand, the analytical theory of dynamical systems also supports deep learning methods in terms of explainable black-box algorithms\cite{Linot2020DeepLT,Liao2021MultiscaleRO,Vogelstein2021SupervisedDR,bernstein2013manifold}. For instance, Vlachas et al. \cite{Vlachas2022MultiscaleSO} present an autoencoder framework with two different time scale updating mechanisms to predict the effective dynamics of some complex systems. 
Manoni et al. \cite{manoni2020effective} develop a manifold learning approach to parametrize biosignal data for generative modeling with Bernstein Polynomial approximations and obtain a latent stochastic model. Lu et al. \cite{lu2017data} construct a data-based predictive reduced model for the Kuramoto–Sivashinsky equation with nonlinear autoregressive moving average as the accurate prediction method. Among these nonlinear dimension reduction techniques, researchers also derive some multiscale regression methods on unknown manifolds from big data, such as brain imaging in biomedical science \cite{Liao2021MultiscaleRO, Vogelstein2021SupervisedDR}. In addition, as data grows out from all kinds of research fields, data driven problems become popular ways to better understand complex real world phenomena through both observation and the first principle of its governing equations. Therefore, all these excellent works inspire us to investigate some innovative algorithms to study the effective systems for data-driven stochastic dynamical systems.

In this paper, we set up our problems as follows. For a given unknown multiscale stochastic dynamical system, suppose we only have access to observation data on an initial short-term period, our goal is to learn effective dynamics with regard to invariant manifold. In fact, the sample paths of the slow-fast stochastic system are concentrated in a neighbourhood of the invariant manifold when the deterministic system admits a uniformly asymptotically stable slow manifold\cite{Berglund2003GeometricSP}. In view of the concentration of the sample paths around invariant manifold, we construct a neural network named Auto-SDE to predict the long-term evolution of the data-driven stochastic dynamical system and approximate the invariant manifold effectively and precisely. Moreover, the reduced dynamics can be achieved by restricting the original system to the invariant manifold.

The main contributions of this paper are:
\begin{itemize}
\item{Learn the stochastic governing law from short-term data which satisfies an unknown stochastic differential dynamical system;}
\item{Develop a general deep learning framework to effectively approximate the invariant slow manifold;}
\item{Evaluate the accuracy of the reduced system in terms of sample trajectories and spatial-temporal distribution.}
\end{itemize}

The remainder of this paper is structured as follows. In Section \ref{PS}, we describe a stochastic problem, define the slow manifold and invariant manifold, and introduce the reduced system. We then construct an algorithm with Auto-SDE network to learn effective reduced dynamics in Section \ref{FW}. In Section \ref{NE}, we validate our framework by numerical experiments of two nonlinear slow-fast stochastic dynamical systems. Finally, we summarize the paper with some conclusions in Section \ref{CO}.

\section{Problem setting}\label{PS}
\noindent 

In this section, we consider the slow-fast stochastic system evolves on two different timescales, separated by a small positive parameter $\varepsilon$ and is perturbed by multidimensional position-dependent noise. Given short-term time series data and the unknown slow-fast stochastic dynamical system 
\begin{equation}\label{SDEMODEL}
\left\{
\begin{array}{l}
dx_t = f(x_t,y_t)dt + \sigma_1F(x_t,y_t)dW_t^1,\\[1em]
dy_t=\frac{1}{\varepsilon}g(x_t,y_t)dt+\frac{\sigma_2}{\sqrt{\varepsilon}}G(x_t,y_t)dW_t^2,\\
\end{array}
\right.
\end{equation}
where $0 < \varepsilon \ll 1 $. $x_t \in \mathbb{R}^{n}$ and $y_t \in \mathbb{R}^{N-n}$ are slow and fast variables respectively. The goal here is to first learn slow invariant manifold from short-term data information and the unknown governing equation (\ref{SDEMODEL}) and then the reduced dynamics on this manifold. 

Let $\mathcal{D}$ be an open set of $\mathbb{R}^{n} \times \mathbb{R}^{N-n}$. The drift functions $f,g$ are twice continuously differentiable and their derivatives are uniformly bounded in $\mathcal{D}$ and similarly for the diffusion functions $F,G$ and their derivatives. The strength of noise in the fast dynamic is chosen to be $\varepsilon^{-1/2}$ to balance the stochastic force and deterministic force. The positive constants $\sigma_1$ and $\sigma_2$ are the intensity of noise acting on the slow and fast components. 

The stochastic processes $W_t^i, i=1,2$ are independent two-side Wiener processes(i.e. Brownian motion) on a probability space $(\Omega,\mathcal{F},\mathbb{P})$. The stochastic integrals with respect to $\{W_t^i\}$ are Ito integrals. The assumptions on the drift coefficients $f$ and $g$ ensure the existence of a continuous version of $(x_t,y_t)$. Hence, we may assume that the path $\omega \rightarrow (x_t(\omega),y_t(\omega))$ is continuous for $\mathbb{P}$-almost all $\omega \in \Omega$.

\subsection{Slow manifold}
\noindent 

We recall the deterministic part in the slow-fast stochastic system \eqref{SDEMODEL}, which is written in the form
\begin{equation}\label{ODE}
\left\{
\begin{array}{l}
dx_t = f(x_t,y_t)dt, \\[1em]
dy_t =\frac{1}{\varepsilon}g(x_t,y_t)dt.\\
\end{array}
\right.
\end{equation}
The particularity of a slow–fast system such as (\ref{ODE}) is that, instead of remaining a system of coupled differential equations, in the limit when $\varepsilon$ tends to zero, it becomes an algebraic differential system, called singularly perturbed system. Setting $t = \varepsilon s$, the system (\ref{ODE}) is converted into a regular perturbation problem as
\begin{equation}\label{ODE1}
\left\{
\begin{array}{l}
dx_s = \varepsilon f(x_s,y_s)ds, \\[1em]
dy_s=g(x_s,y_s)ds.\\
\end{array}
\right.
\end{equation}

The simplest situation occurs when the system (\ref{ODE1}) admits one or several hyperbolic equilibrium points on which $g$ vanishes, while the Jacobian matrix $\partial_y g$ has no eigenvalue on the imaginary axis. Collections of such points define slow manifolds of the system.

\begin{definition}
(Slow manifold)\cite{Berglund2005NoiseInducedPI}. Let $\mathcal{D}_x \subset \mathbb{R}^{n}$ and $\mathcal{D}_y \subset \mathbb{R}^{N-n}$ be connected sets of nonempty interior. Assume that there exists a continuous function $Y$: $\mathcal{D}_x \rightarrow \mathbb{R}^{N-n}$ ($X$: $\mathcal{D}_y \rightarrow \mathbb{R}^{n}$) such that $(x,Y(x)) \in \mathcal{D}$ \ ($(X(y),y) \in \mathcal{D}$) and 
\begin{align}
    g(x,Y(x)) &= 0, x \in \mathcal{D}_x \\ 
    or \ g(X(y),y) &= 0, y \in \mathcal{D}_y.
\end{align}
Then the set $\mathcal{M} = \{(x,y): x \in \mathcal{D}_x, y = Y(x)\}$ ($\mathcal{M} = \{(x,y): y \in \mathcal{D}_y, x = X(y)\}$) is called a slow manifold of the system (\ref{ODE}). Moreover, denote the stability matrix of the associated system at $Y(x)$  (or $y$) by $A(x) = \partial_yg(x,Y(x))$ ($A(X) = \partial_yg(X(y),y)$). The slow manifold $\mathcal{M}$ is called uniformly asymptotically stable if all eigenvalues of $A(x)$($A(X)$) have negative real parts, uniformly bounded away from $0$ for all $x \in \mathcal{D}_x$($y \in \mathcal{D}_y$).
\end{definition}

The slow manifold of the slow-fast system is the manifold where the fast variable can be represented by the slow variable as the scale parameter $\varepsilon \rightarrow 0$. Hence, the dynamics of the slow manifold can be described by the reduced system (or slow system)
\begin{align*}
    &dx_t = f(x_t,{Y}(x_t))dt , \\
   or \ &d{X}(y_t) = f({X}(y_t),y_t)dt.
\end{align*}
The effective system gives a good approximation of dynamical behaviours of the deterministic system. We then come back to the slow-fast stochastic system and introduce the reduced system on the associated invariant manifold.

\subsection{Reduced system on invariant manifold}
\noindent 

For the stochastic system (\ref{SDEMODEL}), the fast variable $y$ is assumed to be exponentially attracted by an invariant manifold if the corresponding deterministic system admits an asymptotically stable slow manifold. Geometric singular perturbation theory implies the existence of the invariant manifold $\mathcal{M}_\varepsilon$. In other words, under suitable conditions, there exists a uniformly asymptotically stable invariant manifold for the stochastic system (\ref{SDEMODEL}). Thus, all orbits from arbitrary points will arrive in the neighbourhood of the invariant manifold and stay close to the invariant manifold as long as the slow dynamics permits.

To construct the invariant manifold, we introduce the following hypotheses with equations (\ref{SDEMODEL}):\\
($\mathcal{H}_1$). The functions satisfying $f \in \mathcal{C}^2(\mathcal{D},\mathbb{R}^{n})$, $g \in \mathcal{C}^2(\mathcal{D},\mathbb{R}^{N-n})$, $F \in \mathcal{C}^1(\mathcal{D},\mathbb{R}^{n} \times \mathbb{R}^{n})$ and $G \in \mathcal{C}^1(\mathcal{D},\mathbb{R}^{N-n} \times \mathbb{R}^{N-n})$, where $\mathcal{D}$ is an open set of $\mathbb{R}^{n} \times \mathbb{R}^{N-n}$. Furthermore, we assume that $f, g, F$ and $G$ are bounded by a constant within $\mathcal{D}$.\\
($\mathcal{H}_2$). With $F = G = 0$, the deterministic system of (\ref{SDEMODEL}) admits a uniformly asymptotically stable slow manifold.\\
($\mathcal{H}_3$). The diffusivity matrix $F(x,y)F(x,y)^T$ is positive definite.
Under these assumptions, Fenichel’s theorem \cite{Berglund2005NoiseInducedPI} ensures the existence of the invariant manifold
\begin{align}
    \mathcal{M}_\varepsilon &=\{(x,y):x \in \mathcal{D}_x, y = \hat{Y}(x)\}, \\ 
    or \ \mathcal{M}_\varepsilon &=\{(x,y):y \in \mathcal{D}_y, x = \hat{X}(y)\},
\end{align}
where $\hat{Y}(x) = Y(x) + O(\varepsilon)$ ($\hat{X}(y) = X(y) + O(\varepsilon)$). They also prove that the sample paths of the stochastic system are concentrated in a neighbourhood of the invariant manifold $\mathcal{M}_\varepsilon$.

Inspired by the concentration of sample paths and deep learning methods, we construct a framework to learn reduced dynamics from data-driven stochastic dynamical systems by approximating the invariant manifold. Due to the curse of dimensionality, we only have access to short-term observations to identify the unknown stochastic differential equations and then predict the long-term dynamics to give the approximation of the invariant manifold. Therefore, the stochastic dynamics can be reduced to the slow systems on the approximated invariant manifold, i.e.
\begin{align*}
    &dx_t = f(x_t,\hat{Y}(x_t))dt + \sigma_1F(x_t,\hat{Y}(x_t))dW_t^1, \\
   or \ &d\hat{X}(y_t) = f(\hat{X}(y_t),y_t)dt + \sigma_1F(\hat{X}(y_t),y_t)dW_t^1.
\end{align*}
The reduced system gives an excellent and effective approximation of the original system. We will illustrate our framework with more details in Section \ref{FW} and show its effectiveness in Section \ref{NE}.

\section{Methods}\label{FW}
\noindent 

In this section, we present a deep learning framework for learning the effective dynamics of the data-driven slow-fast stochastic differential equations. Given access to a black-box simulator of the stochastic process and various initial conditions, we can simulate short-term trajectories as the input to train the framework and learn the effective dynamics. First, we employ an auxiliary estimation network to identify the unknown stochastic differential equations. We then construct a network named Auto-SDE to recursively and effectively predict the trajectories on lower hidden space to approximate the invariant manifold by two key architectures: recurrent neural network and autoencoder. Thus, the reduced dynamics are obtained by time evolution on the invariant manifold. Finally, we demonstrate the loss functions and the unsupervised training algorithms. 

\subsection{Recovery of the slow-fast system}\label{sec3.1}
\noindent 

The autoencoder is often trained by  minimizing the reconstruction loss. The reconstruction data and the input data only partially coincide when it comes to propagating information over time. We use an estimating network to help build up the non-overlapping loss by identifying the stochastic differential equations because the non-overlapping section is out of reconstruction. 

For the stochastic differential equations, the
Kramers–Moyal formula \cite{Schuss2010TheoryAA} introduces the relationship between the short-term data and the coefficients. Specifically, consider the following data-driven stochastic differential equation
\begin{equation*}
    dz_t = f_z(z_t,\theta)dt + \sigma_z(z_t,\theta)dW_t,
\end{equation*}
where $z_t = (x_t,y_t)$ are variables in system (\ref{SDEMODEL}) and $\theta$ is a coefficient set of the parametric orthonormal basis functions, shown in tables for the drift and the diffusion. In this paper, we select polynomials as the basis function set $\mathbf{\Theta}(z)=\{z^n,n\in \mathbb{N}\}$. Then, the following equations hold
\begin{align}\label{KM1}
    f_z(z_t,\theta) &= \lim_{\Delta t \to 0}\mathbb{E}\left[ \frac{(Z_{t+\Delta t}-Z_{t})}{\Delta t} \bigg| Z_t = z_t \right] = \theta^1 \mathbf{\Theta}(z),\\
    \label{KM2}
    \sigma_z^2(z_t,\theta) &= \lim_{\Delta t \to 0}\mathbb{E}\left[ \frac{(Z_{t+\Delta t}-Z_t)^2}{\Delta t} \bigg| Z_t = z_t \right] = \theta^2 \mathbf{\Theta}(z).
\end{align}

We have the outputs of estimation network $f_z(z_t,\theta)$ and $\sigma_z(z_t,\theta)$, whose inputs and weights are represented by the basis functions and coefficients, respectively. Let time series sequence $Z \in \mathbb{R}^{D \times m}$, where $D$ is the space dimension and $m$ is the time window length. We denote $Z = [Z_{t_1}, ..., Z_{t_m}]$ , and at each time step, $Z_{t_i} = [z_{t_i}^1, ..., z_{t_i}^D]^T$.
By the equations (\ref{KM1}) and (\ref{KM2}), the loss functions have the form
\begin{align}
    \mathcal{L}_{drift} &= \frac{1}{(m-1)D}\sum^{m-1}_{i=1}||\frac{Z_{t_{i+1}}-Z_{t_{i}}}{t_{i+1}-t_i} - f_z(Z_{t_{i}})||^2, \\
    \mathcal{L}_{diffusion} &= \frac{1}{(m-1)D}\sum^{m-1}_{i=1}||\frac{(Z_{t_{i+1}}-Z_{t_{i}})^2}{t_{i+1}-t_i} - \sigma_z^2(Z_{t_{i}})||^2.
\end{align}
The estimation networks({Fig.~$\ref{nn}$(a)}) are trained by minimizing the above loss functions and serve as a supplement to Auto-SDE architecture. The basis function set used in Example 1 is $\mathbf{\Theta}(z)=\{1,x,y,x^2,xy,y^2\}$. During training process, $\{\theta^i_j\},i\in \{1,2\},j \in \{1,2,3,4,5,6\}$ will be set to zero if $|\theta^i_j|<0.05$. The final results of $\theta$ are shown in tables.

\subsection{Time evolutionary auto-encoder neural network}
\noindent

Due to the complexity of multiscale systems and high dimensional data, our key idea is to evolve data in a lower dimensional state space. Therefore, the autoencoder framework is adopted to project high-dimensional data into a low dimensional space to effectively reduce computational cost and propagate the reduced data over time in this lower dimensional space.

Beginning with $m$ snapshots from short-term observations $\mathbf{Z} = [z_1, ..., z_m] \in \mathbb{R}^{D \times m},  z = (x,y)$, the fully connected autoencoder can be regarded as a nonlinear dimensionality reduction by projection \cite{Hinton2006ReducingTD}, which consists of a single or multiple-layer encode network. The autoencoder maps the input to a lower dimensional latent space and then decodes it to the original dimension at the output, trained to minimize the reconstruction loss $\mathcal{L}_{AE}$.

                                                                            
To propagate reduced lower dimensional data over time, the recurrent neural network (RNN) is a natural choice, as it considers a sequence of data as input, updating the hidden state with the current state and the prior hidden memory. 
However, it is difficult to train RNN for long-term memory. The major challenge is that gradients rise exponentially or decay exponentially when they are backpropagated through each time step. The gradients will either vanish or explode over several time steps. To solve the vanishing or exploding gradient problem, the gated RNN, which includes the long short-term memory network (LSTM) and the gated recurrent unit network, allows gradients of the loss function to backpropagate across many time-steps by using additional routes to make the required parameter update. Here, we embed LSTM units inside the autoencoder framework to predict discrete sequential data from short-term trajectories over time.

Considering the need of long-term evolution on low dimensional space, here we propose our model \textbf{Auto-SDE} network. The Auto-SDE uses autoencoder framework equipped with the LSTM units which contribute to predict long-term dynamics from short-term observations over low dimensional latent space. The benefit from this network is that it overcomes the curse of dimensionality in complex and multiscale systems as well as generating accurate long-term prediction in the original high dimensional space. The invariant manifold are then approximated by training Auto-SDE network recursively over time. And this training process needs a trick to construct an effective loss function $\mathcal{L}_{SDE}$ based on discretized stochastic differential equation.

\subsection{Loss function and training algorithm}
\noindent 

\begin{figure}[htp] 
    \centering
    \includegraphics[width=15cm]{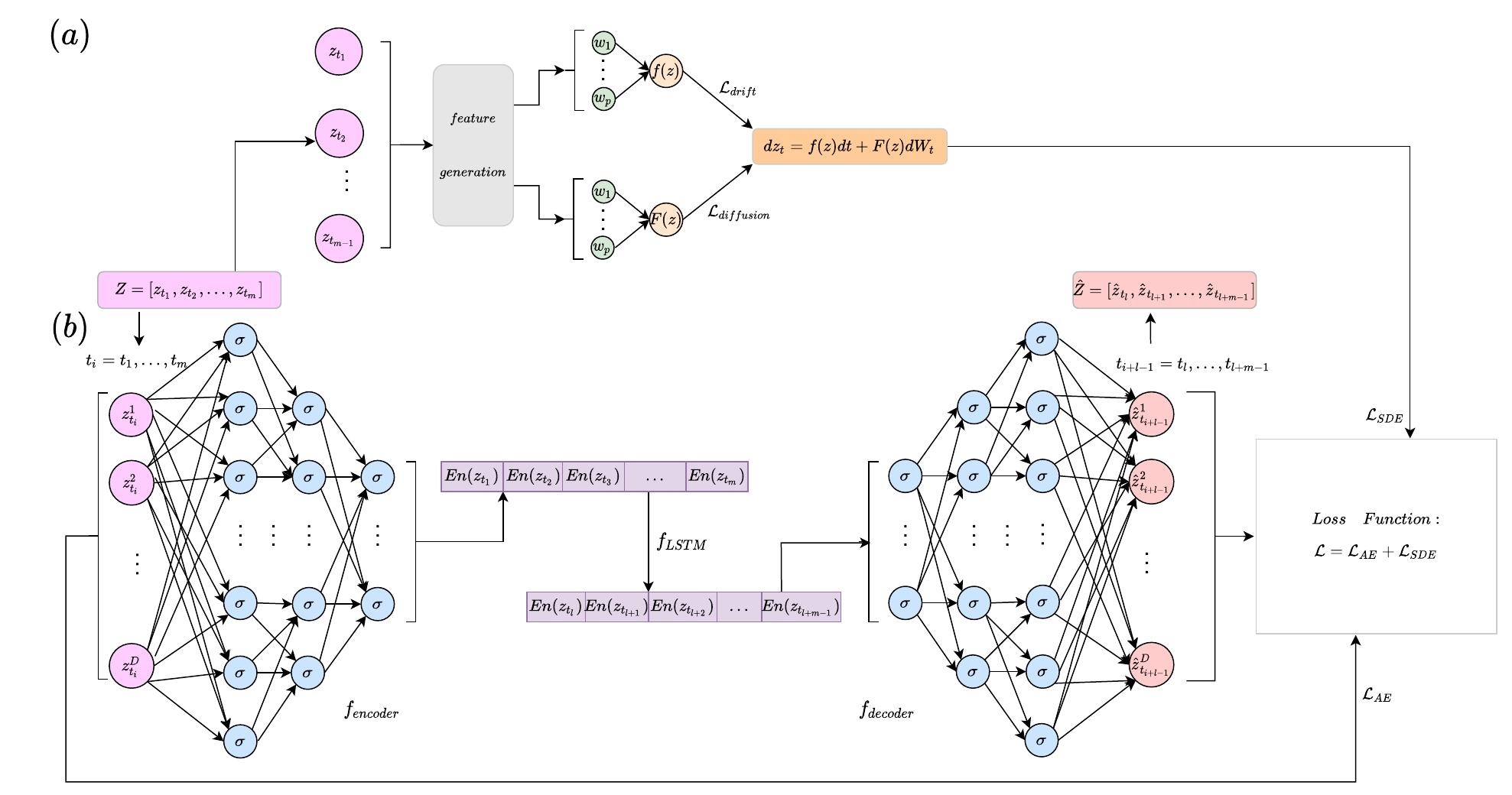}
    \caption{A schematic of our framework and the workflow during the training phase. (a) The auxiliary estimation network to identify the stochastic differential equations. (b) Auto-SDE consists of a 3-layer encoder, a LSTM layer and a 3-layer decoder. The Auto-SDE network adopts overlapping technique with reconstruction and stochastic differential equations to predict time evolution on the lower dimensional representation effectively.}
    \label{nn}
\end{figure}

Due to the structure of our neural network, we seek to construct a loss function consisting of two parts: the reconstruction loss $\mathcal{L}_{AE}$ from the autoencoder for the overlapping part between the input and the output, and the stochastic differential equation loss $\mathcal{L}_{SDE}$ for the non-overlapping part. 
Note that, the stochastic differential equation loss $\mathcal{L}_{SDE}$ is computed by part of the output of the decoder and the identification of the stochastic differential equations from the estimation network, together with part of the input of the encoder. 
We would like to recursively train the model to accurately and efficiently propagate time series data in the lower space. The framework is shown in {Fig.~$\ref{nn}$}.\\

Recall our input data $Z = [Z_{t_1}, ..., Z_{t_m}]$, where $ Z_{t_i} = [z_{t_i}^1, ..., z_{t_i}^D]^T$ and the output $\hat{Z} = [\hat{Z}_{t_l}, ..., \hat{Z}_{t_{l+m-1}}]$ for arbitrary $ 1 < l \leq m$, where $l-1$ is number of the forward prediction steps, $[\hat{Z}_{t_l}, ..., \hat{Z}_{t_{m}}]$ is the reconstruction of $[Z_{t_l}, ..., Z_{t_m}]$ and $[\hat{Z}_{t_{m+1}}, ..., \hat{Z}_{t_{l+m-1}}]$ is the prediction state. In other words, we predict the states in $[t_{m+1},...,t_{l+m-1}]$ by the states in $[t_{1},...,t_{m}]$. The sign $D$ represents the dimension of the data. Let $Z_{SDE}$ denotes the values of $[Z_{t_{m+1}}, ..., Z_{t_{l+m-1}}]$, which can be obtained from Euler-Maruyama approximation of the estimated stochastic differential equation given some initial data points. The reconstruction loss $\mathcal{L}_{AE}$ and the stochastic differential equation loss $\mathcal{L}_{SDE}$ are defined as
\begin{equation}\label{losae}
    \mathcal{L}_{AE} = \mathcal{L}_{AE}(\hat{Z},Z)
    = \frac{1}{(m-l+1)D}\sum^{m}_{i=l}||\hat{Z}_{t_i}-Z_{t_i}||^2,
\end{equation}
and
\begin{equation}\label{lossde}
    \mathcal{L}_{SDE} = \mathcal{L}_{SDE}(\hat{Z},Z_{SDE})
    = \frac{1}{(l-1)D}\sum^{l+m-1}_{i=m+1}||\hat{Z}_{t_i}-Z_{t_i}||^2.
\end{equation}
Therefore, the loss function of Auto-SDE is written as
\begin{equation}\label{losfun}
    \begin{split}
    \mathcal{L}(\hat{Z},Z,Z_{SDE}) &= \mathcal{L}_{AE} + \mathcal{L}_{SDE}\\
    &= \frac{1}{(m-l+1)D}\sum^{m}_{i=l}||\hat{Z}_{t_i}-Z_{t_i}||^2 + \frac{1}{(l-1)D}\sum^{l+m-1}_{i=m+1}||\hat{Z}_{t_i}-Z_{t_i}||^2.
    \end{split}
\end{equation}

\begin{algorithm}
    \caption{Training Algorithm of Auto-SDE.}
    \label{alg1}
    \KwIn{Initial dataset $\mathcal{Z} \in \mathbb{R}^{D \times m}$, number of training epochs $N_e$, bath size $N_b$, and estimation parameter set $\theta$.}
    \KwOut{The approximated invariant manifold and the reduced dynamics.}
    \BlankLine
    Initialize the model parameters ${\vartheta_{mol}}$ randomly;
    
    $\mathcal{X}_{in} = \mathcal{Z} \in \mathbb{R}^{D \times m}$, $\mathcal{X}_{in} := [\mathcal{X}_{in}^1, ..., \mathcal{X}_{in}^m]$;
    
    \While{$dist(\mathcal{X}_{in}^{m-1},\mathcal{X}_{in}^m)$ not converges}{
    
    \For{$i \in \{1, ..., N_e\}$}{
    Randomly sample batch from dataset: $\mathcal{Z}_b^i \subset \mathcal{X}_{in}, \mathcal{Z}_b^i \in \mathbb{R}^{N_b \times m}$;
    
    Encoder forward pass: $\mathcal{H}_b^i \gets f_{encoder}(\mathcal{Z}_b^i)$;
    
    Overlapping prediction forward pass: $\hat{\mathcal{H}}_b^i  \gets f_{LSTM}(\mathcal{H}_b^i)$;
    
    Decoder forward pass: $\hat{\mathcal{Z}}_b^i \gets f_{decoder}(\hat{\mathcal{H}}_b^i)$;
    
    Using $\mathcal{Z}_b^i$, $\hat{\mathcal{Z}}_b^i$ and $\mathcal{Z}_{SDE,b}^i = f_{\theta}(\mathcal{Z}_b^i)$ to calculate approximate gradient of the loss function (\ref{losfun}) and update the model parameters ${\vartheta_{mol}}$;
    }
    
    $\mathcal{X}_{in} = \hat{\mathcal{Z}} = f_{\vartheta_{mol}}(\mathcal{X}_{in})$;
    }
    
    Approximate the invariant manifold by $\mathcal{X}_{in}^m$;
    
    Learn the reduced dynamics by the invariant manifold and the data-driven stochastic differential equation(s). 
   
\end{algorithm}

\section{Numerical experiments}\label{NE}
\noindent 

We apply the methods described in the previous sections on examples to illustrate the validity of deep learning approaches to learn effective dynamics. The approximation of the invariant manifold and reduced system are obtained in our framework. We assume that the differential equations are unknown. We will demonstrate our framework with more details. To solve the optimization problem, we use the ADAM optimizer with gradient. The gradient computation optimizes both loss functions for reconstruction and data-driven stochastic differential equations. In all numerical examples, we randomly generate the initial conditions for the system. For the ADAM method, we choose the learning rate as $10^{-3}$.

\begin{example}\label{2dexam}
Consider a simple nonlinear slow-fast stochastic system satisfying
\begin{equation}\label{ex1}
\left\{
\begin{array}{l}
dx = (x-xy)dt +\sigma_1dW_t^1,\\
dy = -\frac{1}{\varepsilon}(y-\frac{1}{4}x^2)dt+\frac{\sigma_2}{\sqrt{\varepsilon}}dW_t^2,\\
\end{array}
\right.
\end{equation}
where $\sigma_1  = 1$, $\sigma_2  = 0.2$ and $\varepsilon = 0.01$. To generate the data, we use a time step of $\Delta t = 0.001$ with $x(0) \sim Uni(-5,5)$ and $y(0) \sim Uni(-4,4)$. We aim to use Auto-SDE to learn an effective reduced-order model for the multiscale system based on the short-term simulation data. Once the effective model is built, we can adopt it as a surrogate model to extrapolate the reduced dynamics for a long time. We simulate $1200$ samples with different initial conditions and choose $[t_0, t_m]=[0, 0.01]$ to create a short-term dataset as the input to learn the reduced dynamics. 
\end{example} 

With the given input dataset on $[0, 0.01]$ and method in Section \ref{sec3.1}, the estimation for the coefficients of the original system is shown in Table \ref{2B esti}. 
The sign 'Learnt' represents the identification of the coefficients of the basis functions and 'True' means the actual functions.
The diffusion term of fast variable is not estimated very well because of the large drift coefficients. But this has little impact on long term prediction.\\

\begin{table}
\centering
\caption{The drift and the diffusion}
\label{2B esti}
\begin{tabular}{c|cc|cc|cc|cc}
\toprule  
\multirow{2}*{basis}& \multicolumn{2}{c|}{\textbf{$x$ drift}}& \multicolumn{2}{c|}{\textbf{$x$ diffusion}} &\multicolumn{2}{c|}{\textbf{$y$ drift}}& \multicolumn{2}{c}{\textbf{$y$ diffusion}} \\
\cline{2-9}
& Learnt & True& Learnt & True & Learnt & True& Learnt & True\\
\midrule  
1 & 0 & 0 & 1.0066 & 1 & 0 & 0 & 3.2442 & 2\\
$x$ & 1.0000 & 1 & 0 & 0 & 0 & 0 & 0 & 0\\
$y$ & 0 & 0 & 0 & 0 & -99.6987 & -100 & 0 & 0\\
$x^2$ & 0 & 0 & 0 & 0 & 25.1285 & 25 & 0 & 0\\
$xy$ & -1.0243 & -1 & 0 & 0 & 0 & 0 & 0 & 0\\
$y^2$ & 0 & 0 & 0 & 0 & 0 & 0 & 0 & 0\\
\bottomrule 
\end{tabular}
\end{table}

{\textbf{The invariant manifold}}

\begin{figure}[htp]
    \centering
    \includegraphics[width=8cm]{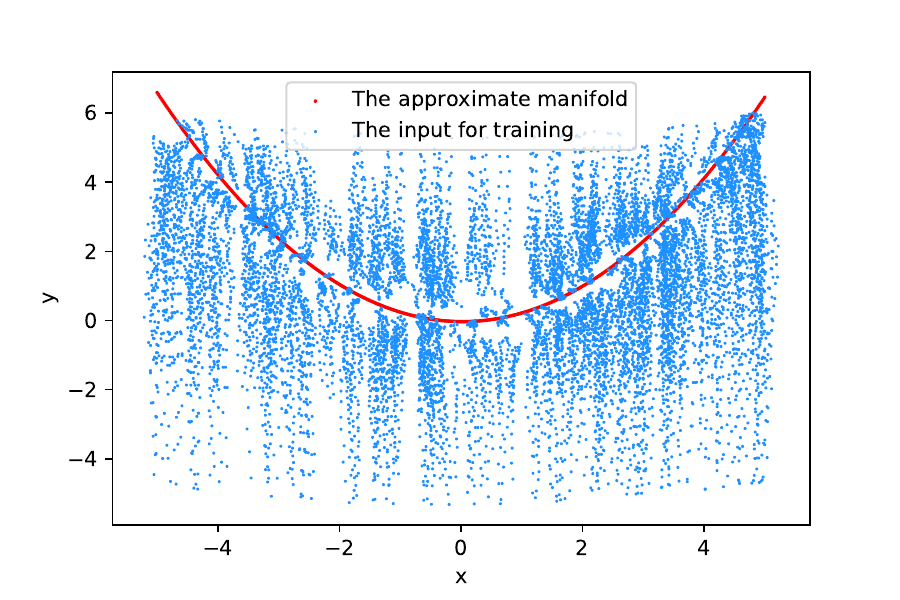}
    \caption{The input dataset (blue dot) and the approximated manifold (red curve).}
    \label{2B mld}
\end{figure}

\begin{figure}[htp]
    \centering
    \includegraphics[width=15cm]{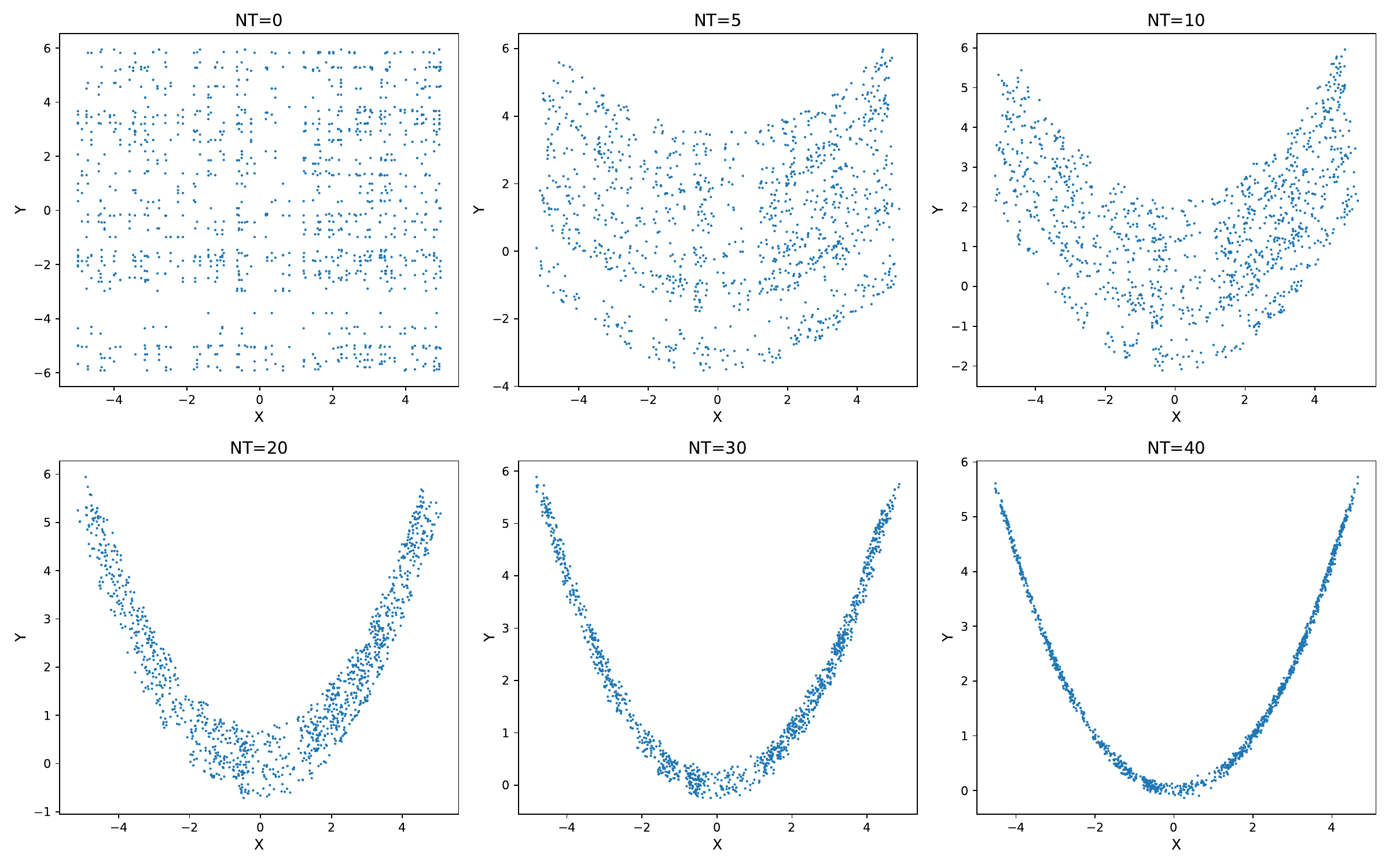}
    \caption{The snapshots of 1200 initial points evolving over time. The mark $NT$ represents the Nth time step with step size $\Delta t =0.001$.}
    \label{2B st}
\end{figure}

The input dataset contains 10 time steps and the approximated manifold are plotted in {Fig.~$\ref{2B mld}$}. To better explain how well Auto-SDE approximate the invariant manifold, {Fig.~$\ref{2B st}$} shows the snapshots of sample points at different time steps. The snapshots at $NT = 0,1,...,10$ are known from the short-term simulation data. When $NT \geq 11$, the snapshots of positions are effectively predicted by Auto-SDE. We select $NT = 0,5,10,20,30,40$ to show the evolution of trajectories in distribution. The network keeps training and predicting recursively until the distribution of positions $x$ and $y$ at some time step is similar to that of the former time step. That means the sample paths are on or close to the invariant manifold and the original system can be reduced to the slow system via the invariant manifold. Specifically, the mean absolute errors between two probability densities of adjacent snapshots are $0.016115$, $0.012667$, $0.007844$ for $NT = 10$ and $20$, $NT = 20$ and $30$, $NT = 30$ and $40$ separately. Moreover, the explicit expression of the approximated invariant manifold calculated by the snapshot data at $NT = 40$ can be written as $\hat{Y}(x) = -0.0324 -0.0134x + 0.2622x^2$.\\

{\textbf{The reduced dynamics}}

\begin{figure}[htp]
    \centering
    \includegraphics[width=8cm]{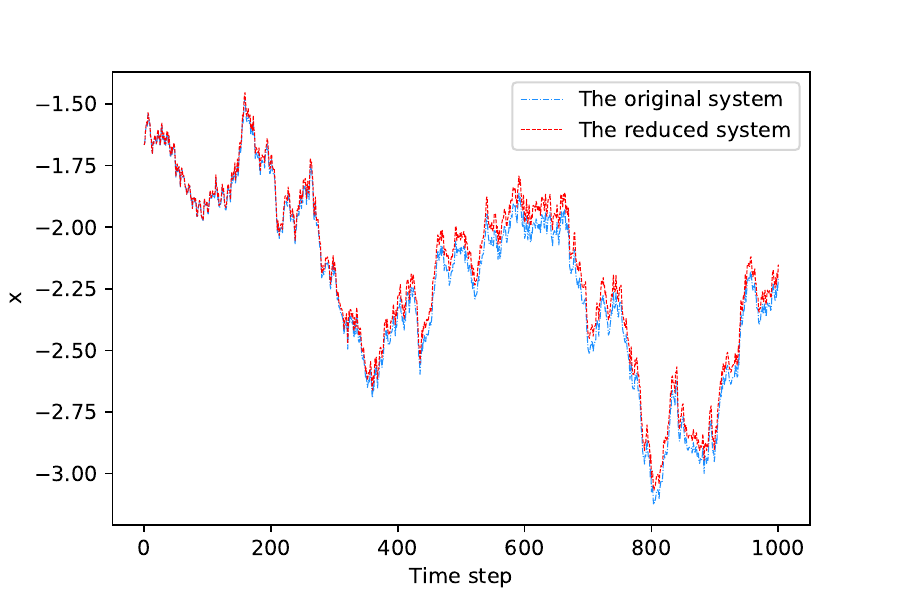}
    \caption{The trajectories generated by the approximated reduced system of our network and the slow variable of the original system. The starting point is on the approximated manifold.}
    \label{2B reds}
\end{figure}

\begin{figure}[htp]
    \centering
    \includegraphics[width=15cm]{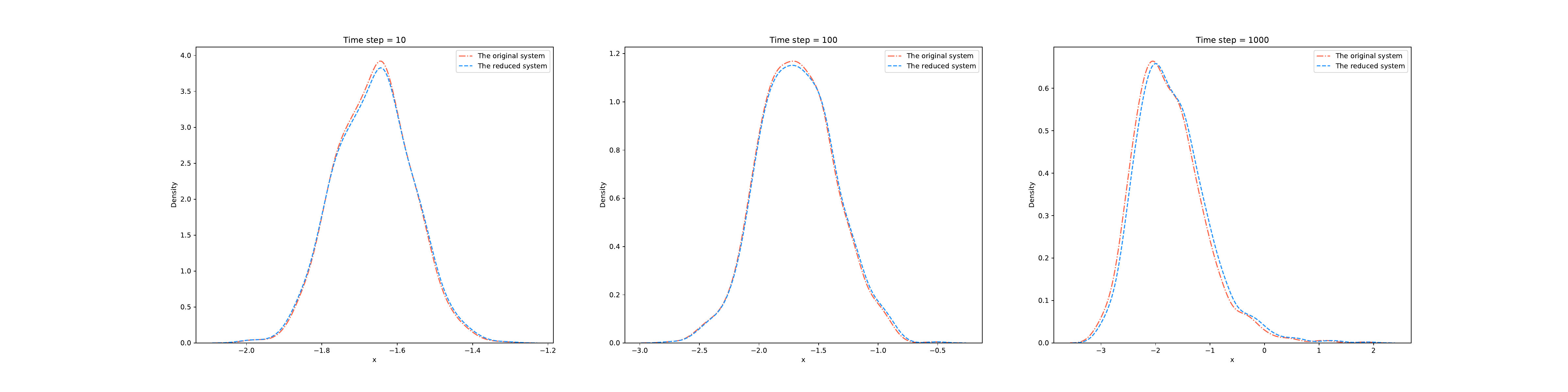}
    \caption{The distribution of 1000 trajectories generated by the approximated reduced system of our network and the slow variable of the original system. From left to right, the value of $NT$ is equal to 10, 100, or 1000.}
    \label{2B reo}
\end{figure}

From the invariant manifold and the identified stochastic differential equations, we can obtain the reduced system. To verify that the reduced system gives a good approximation of the original system, 1000 samples from the same initial condition on the approximated manifold are generated. From the sample path point of view, {Fig.~$\ref{2B reds}$} shows a trajectory generated by our approximation of the reduced system, compared with that of the slow variable in the original system. Due to the stochastic noise, the different time steps and random initial conditions, the trajectory in {Fig.~$\ref{2B reds}$} could change, while the trajectory from our reduced system can always track the original one accurately. 
Furthermore, the distributions of 1000 samples at different time steps of the reduced dynamics are also compared with those of the original slow variable in 
{Fig.~$\ref{2B reo}$}. The reduced dynamics learned by Auto-SDE faithfully reproduces the slow dynamics of the original system, so we can learn multiscale dynamical systems effectively and sufficiently.

\begin{figure}[htp]
    \centering
    \includegraphics[width=12cm]{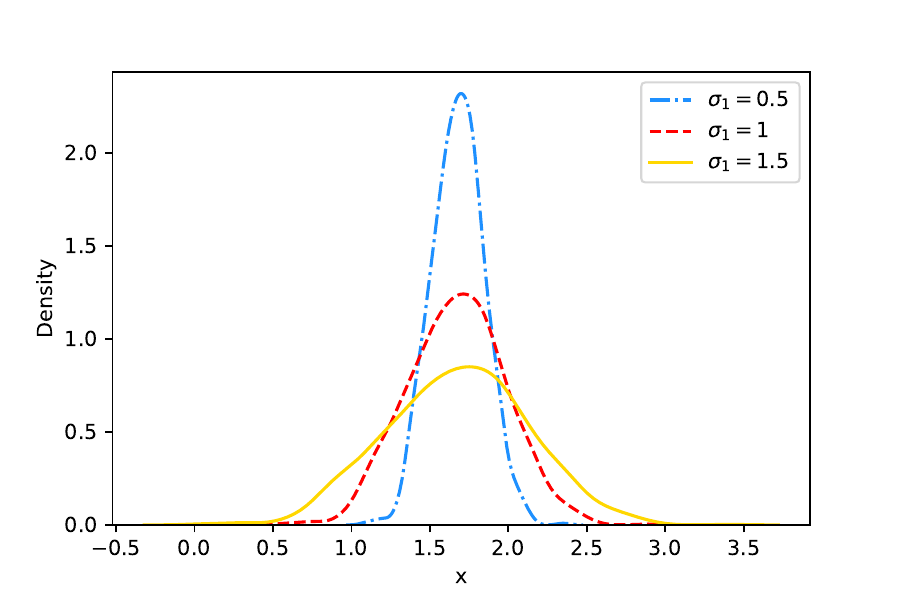}
    \caption{The distribution of 1000 trajectories generated by the approximated reduced systems}. The starting point is the same point near the approximated manifold. The noise density equals to 0.5, 1 and 1.5.
    \label{2B sig}
\end{figure}

We also check the effect of noise strength on the approximated reduced dynamics. The slow variable $x$ follows a Gaussian distribution with regard to $f$, $\sigma_1$ and $t$ since the reduced system of (\ref{ex1}) customarily has the form $dx = f(x,\hat{Y}(x))dt +\sigma_1dW_t^1$. If the noise intensity $\sigma_1$ changes, the distribution of the reduced systems will be different due to the variance. The 1000 sample paths are generated from the same initial condition for three equations with time step $dt = 0.001$ and different $\sigma_1 = 0.5, 1, 1.5$. {Fig.~ $\ref{2B sig}$} displays their distributions at the same time step $NT = 100$. It shows that the distribution of the reduced system with $\sigma_1=0.5$ has the highest peak value and the distribution gently disperses as $\sigma_1$ increases. It is comprehensible since as the noise intensity increases, so does the variance. Then the states are dispersed and the distribution is not concentrated.

\begin{example}
We consider the three-dimensional nonlinear system given by
\begin{equation}\label{ex2}
\left\{
\begin{array}{l}
dx_1 = (x_1+y-\frac{x_2}{2})dt+\sigma_1 dW_t^1,\\
dx_2 = (x_2+y^2-x_1)dt+\sigma_2 dW_t^2,\\
dy = -\frac{1}{\varepsilon}(y-\frac{1}{4}x_1)dt+\frac{\sigma_3}{\sqrt{\varepsilon}}dW_t^3.\\
\end{array}
\right.
\end{equation}
Here, $\sigma_1 = 1$, $\sigma_2 = 2$, $\sigma_3 = 0.2$ and $\varepsilon = 0.01$. We model the stochastic process with initial conditions on a bounded domain $x_1(0) \sim Uni(-4,4)$, $x_2(0) \sim Uni(-4,4)$ and $y(0) \sim Uni(-4,4)$. We construct and train Auto-SDE to extrapolate the reduced dynamics for a long-term period. To obtain the input dataset, we solve the system by the Euler-Maruyama scheme with step size $dt = 0.001$ for a short-term interval $[t_0,t_m] = [0,0.01]$. 
\end{example}

Note that we now have access to a short-term period of observations denoted by $Z_t, t \in [0,0.01]$. The approximated manifold and the reduced dynamics are obtained by Auto-SDE. The estimation for the coefficients of the original system is shown in Table $\ref{3B esti}$.\\
\begin{table}
\caption{The drift and the diffusion}
\label{3B esti}
\centering
\begin{tabular}{c|cc|cc|cc|cc|cc|cc}
\toprule  
\multirow{2}*{basis}& \multicolumn{2}{c|}{\textbf{$x_1$ drift}}& \multicolumn{2}{c|}{\textbf{$x_1$ diffusion}}& \multicolumn{2}{c|}{\textbf{$x_2$ drift}}& \multicolumn{2}{c|}{\textbf{$x_2$ diffusion}}& \multicolumn{2}{c|}{\textbf{$y$ drift}}& \multicolumn{2}{c}{\textbf{$y$ diffusion}} \\
\cline{2-13}
& Learnt & True & Learnt & True& Learnt & True& Learnt & True& Learnt & True& Learnt & True\\
\midrule  
1& 0 & 0 & 1.0044 & 1 & 0 & 0 & 2.0057 & 2 & 0 & 0 & 2.6082 & 2\\
$x_1$& 1.0497 & 1 & 0 & 0 & -1.1081 & -1 & 0 & 0& 25.3413 & 25 & 0 & 0\\
$x_2$& -0.5039 & -0.5 & 0 & 0 & 0.9708 & 1 & 0 & 0& 0 & 0 & 0 & 0\\
$y$& 0.9923 & 1 & 0 & 0 & 0 & 0 & 0 & 0& -99.6571 & -100 & 0 & 0\\
$x_1^2$& 0 & 0 & 0 & 0 & 0 & 0 & 0 & 0& 0 & 0 & 0 & 0\\
$x_1x_2$& 0 & 0 & 0 & 0 & 0 & 0 & 0 & 0& 0 & 0 & 0 & 0\\
$x_1y$& 0 & 0 & 0 & 0 & 0 & 0 & 0 & 0& 0 & 0 & 0 & 0\\
$x_2^2$& 0 & 0 & 0 & 0 & 0 & 0 & 0 & 0& 0 & 0 & 0 & 0\\
$x_2y$& 0 & 0 & 0 & 0 & 0 & 0 & 0 & 0& 0 & 0 & 0 & 0\\
$y^2$& 0 & 0 & 0 & 0 & 0.9782 & 1 & 0 & 0& 0 & 0 & 0 & 0\\
\bottomrule 
\end{tabular}
\end{table}

{\textbf{The invariant manifold}}

\begin{figure}[htp]
    \centering
    \includegraphics[width=15cm]{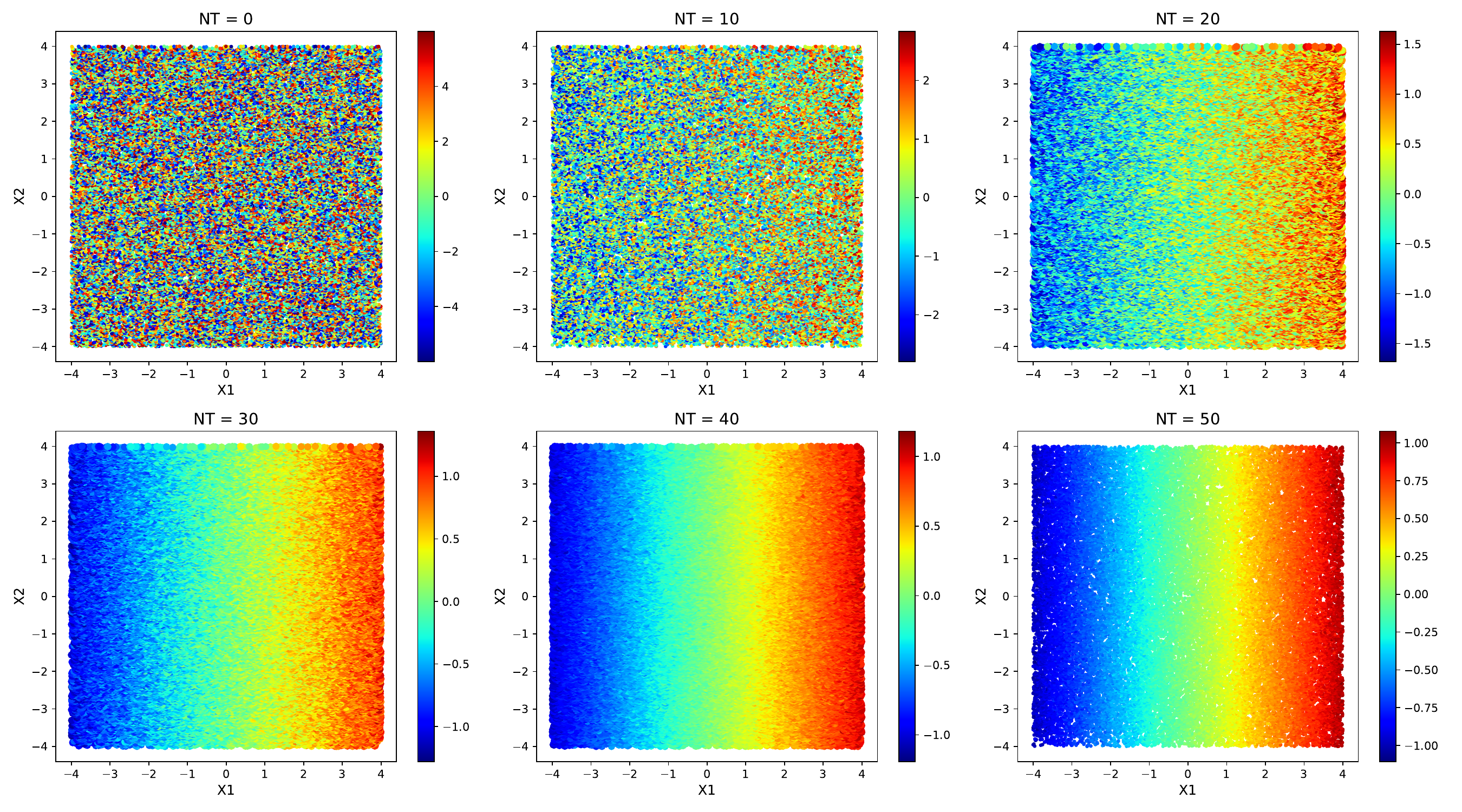}
    \caption{The snapshots of 40000 initial points evolving over time. The time step size is $dt = 0.001$ and $NT$ denotes the $Nth$ time step. We know the data from $NT = 0$ to $NT = 10$ and predict data at $NT > 10$.}
    \label{3B snap}
\end{figure}

We project the three-dimensional coordinates $(x_1,x_2,y)$ to the plane $x_1Ox_2$ to better understand the input dataset and the prediction dynamics in {Fig.~$\ref{3B snap}$}. Similar to Example \ref{2dexam}, we present the evolution of trajectories in distribution. {Fig.~$\ref{3B snap}$} shows that the distribution of values of the fast variable $y$ becomes much smoother at $NT = 50$ and is close to those of the former time steps. The mean absolute errors of $y$ values between two adjacent snapshots are $0.206738$, $0.070837$, $0.026578$, $0.015885$ for $NT = 10$ and $20$, $NT = 20$ and $30$, $NT = 30$ and $40$, $NT = 40$ and $50$ separately. That means the sample paths are on or near the invariant manifold and the original system can be reduced based on the invariant manifold. The approximated invariant manifold, which has the form $\hat{Y}(x_1,x_2)=-0.0009-0.2457x_1-0.0012x_2$, is calculated by the snapshot at $NT = 50$.\\

{\textbf{The reduced dynamics}}

\begin{figure}[htp]
    \centering
    \includegraphics[width=8cm]{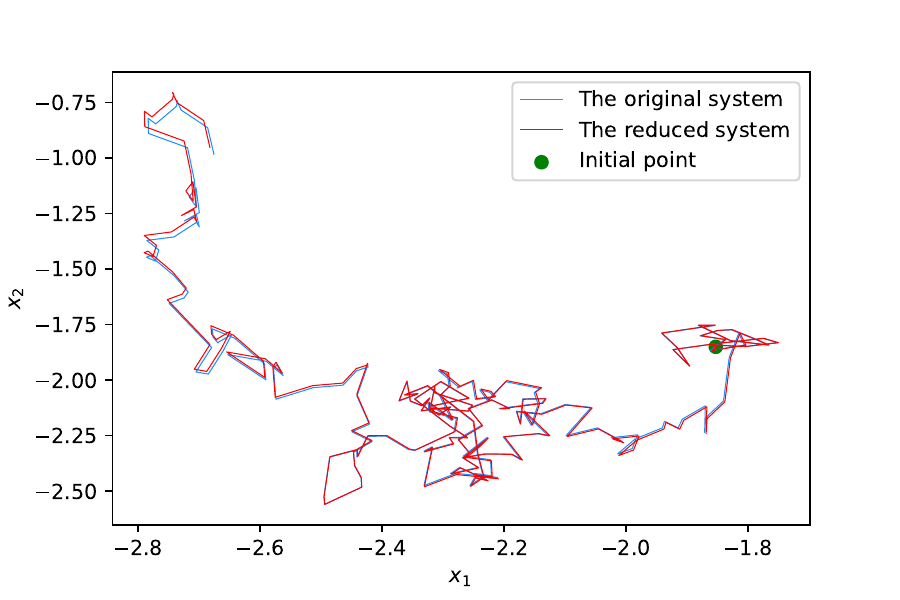}
    \caption{The trajectories generated by the approximated reduced system of our network and the slow variables of the original system. The starting point is on the approximated manifold.}
    \label{3B reds}
\end{figure}

\begin{figure}[htp]
    \centering
    \includegraphics[width=15cm]{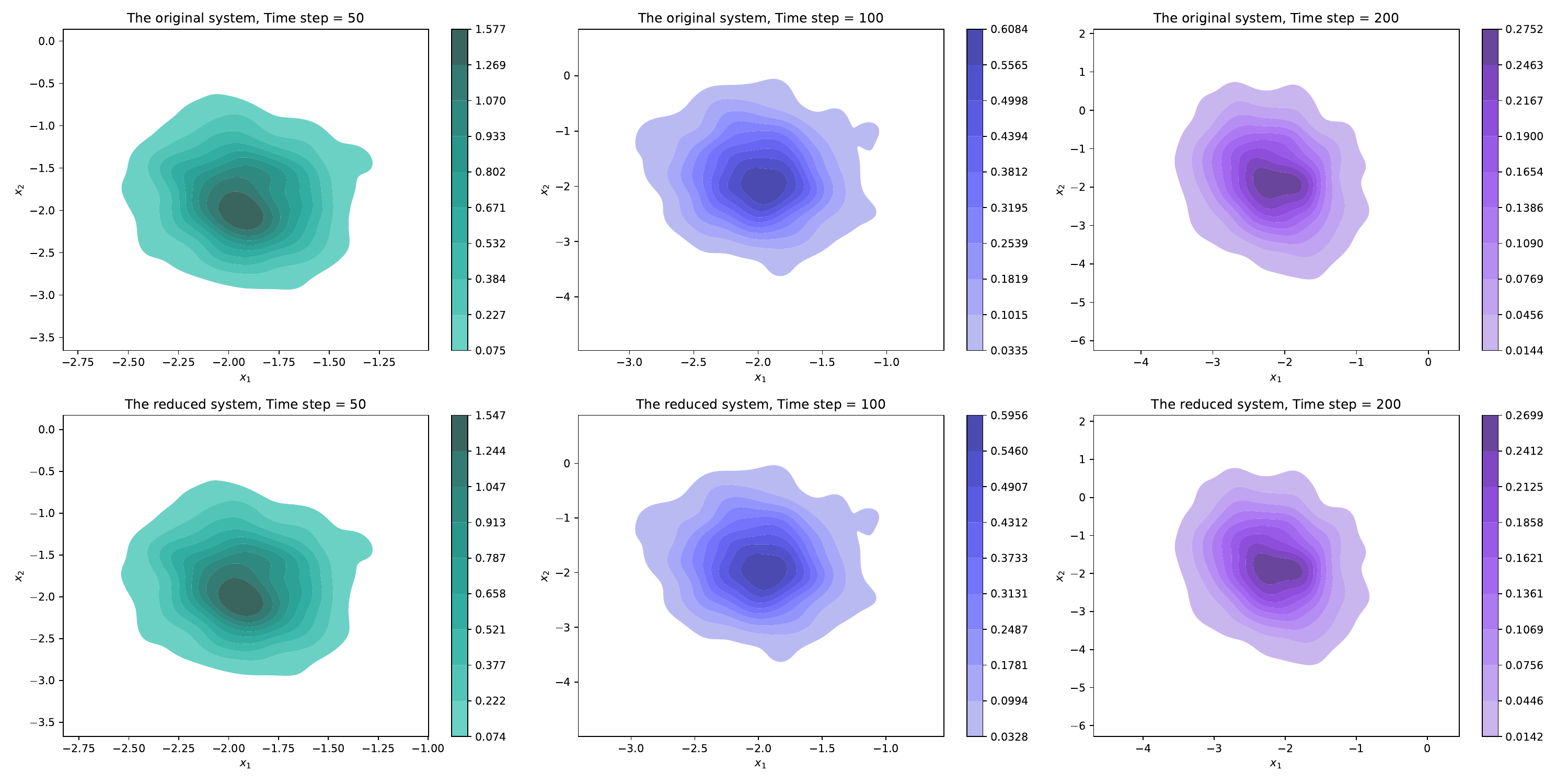}
    \caption{The distribution of 1000 trajectories generated by the approximated reduced system of our network and the slow variable of the original system. From left to right, the value of $NT$ is equal to 50, 100, or 200.}
    \label{3B reo}
\end{figure}

\begin{figure}[htp]
    \centering
    \includegraphics[width=15cm]{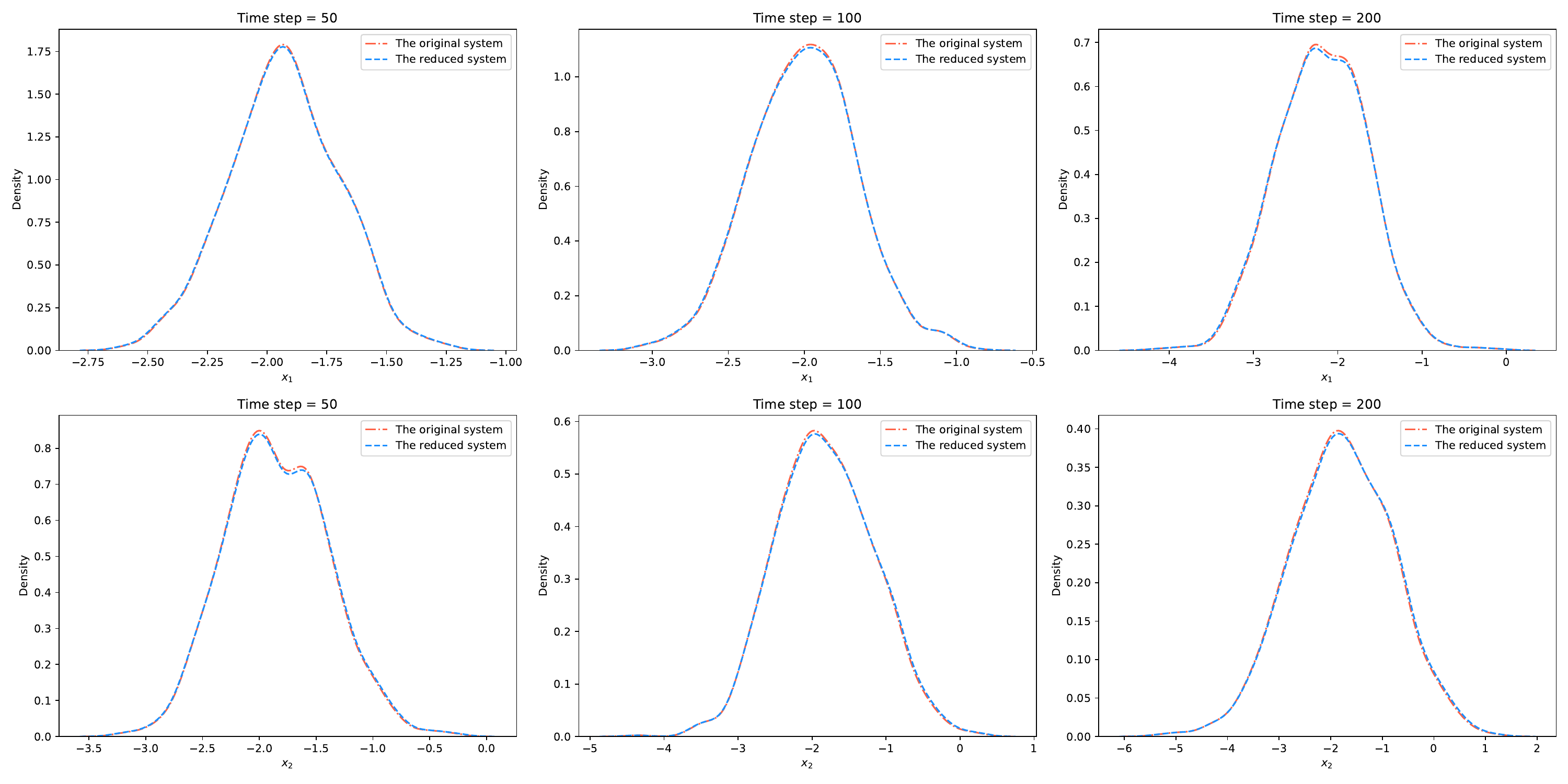}
    \caption{The distribution of 1000 trajectories generated by the approximated reduced system of our network and the slow variable of the original system, shown by $x_1$ and $x_2$ separately. From left to right, the value of $NT$ is 50, 100, or 200.}
    \label{3B reos}
\end{figure}

We then obtain the reduced system by projecting the identified stochastic differential equations onto the invariant manifold. For validation, 1000 samples from the same initial condition on the approximated manifold are generated. {Fig.~$\ref{3B reds}$} shows our approximation of the reduced system, compared with the slow variables in the original system. The trajectory of the reduced system varies due to the random noise, however, it is always consistent with that of the original slow variables {Fig.~$\ref{3B reds}$}. Furthermore, the distributions of 1000 samples at different time steps of the reduced dynamics are compared with those of the original slow variable in {Fig.~$\ref{3B reo}$ } and {Fig.~$\ref{3B reos}$ } by projecting to slow variable coordinates. The reduced dynamics by Auto-SDE correctly recreate the dynamics of the original system on slow variables, allowing us to learn multiscale dynamical systems efficiently.

\begin{figure}[htp]
    \centering
    \includegraphics[width=15cm]{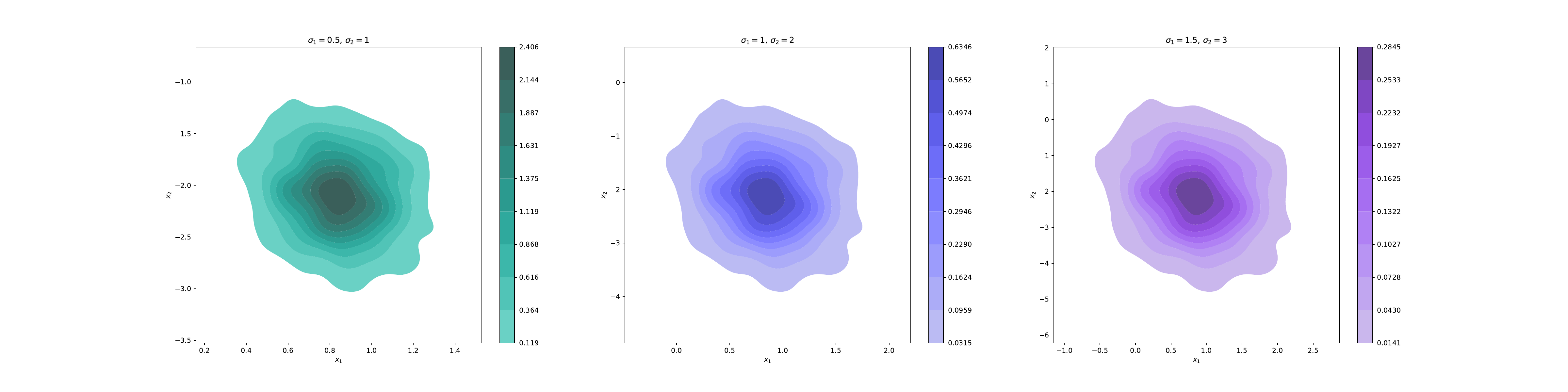}
    \caption{The distribution of 1000 trajectories generated by the approximated reduced systems}. The starting point is the same point near the approximated manifold. The noise density equals to $(0.5, 1), (1, 2)$, and $(1.5, 3)$.
    \label{3B sig}
\end{figure}

\begin{figure}[htp]
    \centering
    \includegraphics[width=15cm]{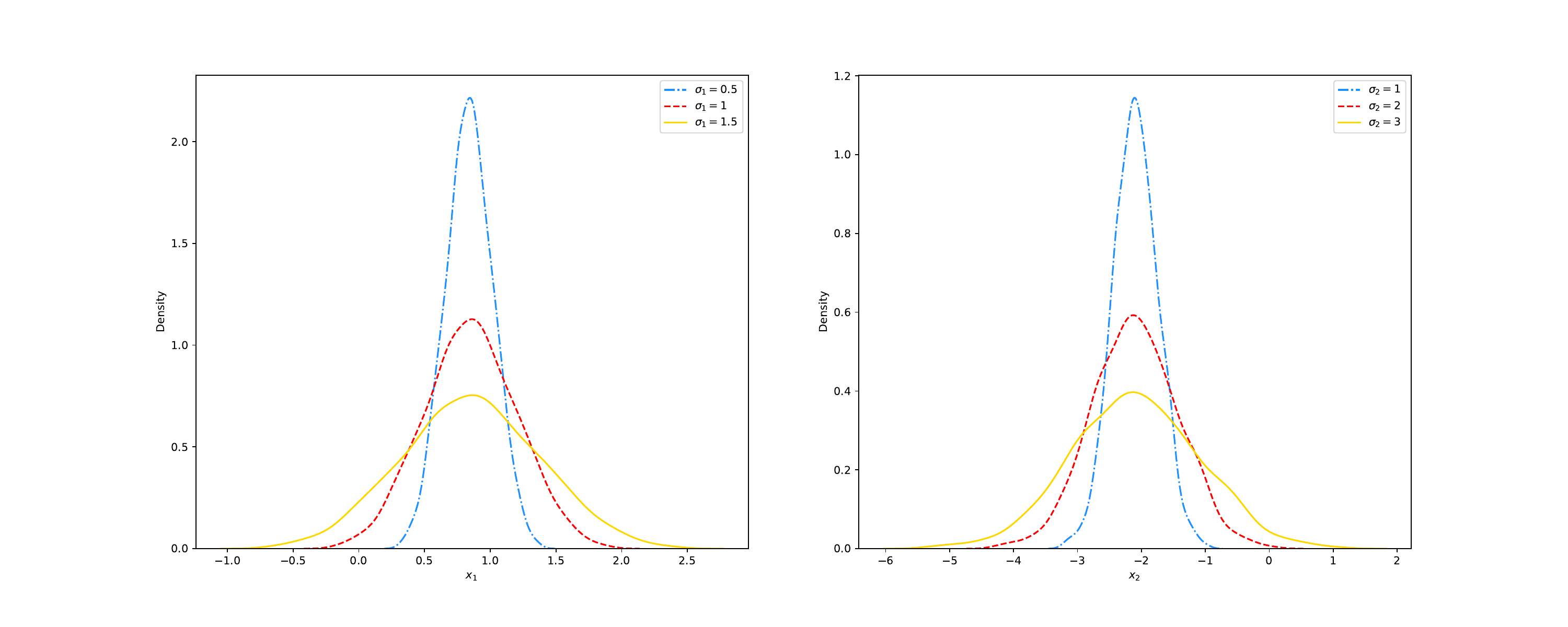}
    \caption{The distribution of 1000 trajectories generated by the approximated reduced systems},  shown by $x_1$ and $x_2$ separately. The starting point is the same point near the approximated manifold. The noise density equals to $(0.5, 1), (1, 2)$, and $(1.5, 3)$.
    \label{3B sigs}
\end{figure}

The influence of the noise intensity on the approximated reduced dynamics are also verified. The slow variables follow Gaussian distribution, so we change the noise intensities $\sigma_1$ and $\sigma_2$ to show the transformation. The 1000 sample paths are generated from the same initial condition for three equations with $dt = 0.1$ and $(\sigma_1, \sigma_2) = (0.5, 1), (1, 2)$, and $(1.5, 3)$. {Fig.~$\ref{3B sig}$} and {Fig.~$\ref{3B sigs}$} show the distribution of 1000 states at $NT=100$. They show that the distribution of the reduced system with $(\sigma_1, \sigma_2) = (0.5, 1)$ gives the sharpest peak value and the distribution slopes gently with the increase of $(\sigma_1, \sigma_2)$. It is understandable that as the noise intensity increases, the system is much more affected and has more dispersed distribution. 

\begin{example}
We consider the stochastic Van der Pol system given by
\begin{equation}\label{ex3}
\left\{
\begin{array}{l}
dx = -ydt+\sigma_1 dW_t^1,\\
dy = \frac{1}{\varepsilon}(y-\frac{1}{3}y^3+x)dt+\frac{\sigma_2}{\sqrt{\varepsilon}}dW_t^2.\\
\end{array}
\right.
\end{equation}
Here, $\sigma_1 = 1$, $\sigma_2 = 0.2$ and $\varepsilon = 0.04$. We model the stochastic process with initial conditions on a bounded domain $x(0) \sim Uni(-2,2)$ and $y(0) \sim Uni(-2,2)$. We construct and train Auto-SDE to extrapolate the effective dynamics for a long-term period. To obtain the input dataset, we solve the system by the Euler-Maruyama scheme with step size $dt = 0.01$ for a short-term interval $[t_0,t_m] = [0,0.1]$ and $2400$ trajectories are generated for training and predicting.
\end{example}

Similarly, assuming we have access to a short-term period of observations denoted by $Z_t, t \in [0,0.1]$. The approximated manifold is obtained by Auto-SDE. The estimation for the coefficients of the original system is shown in Table $\ref{SVDP esti}$.\\
\begin{table}
\centering
\caption{The drift and the diffusion}
\label{SVDP esti}
\begin{tabular}{c|cc|cc|cc|cc}
\toprule  
\multirow{2}*{basis}& \multicolumn{2}{c|}{\textbf{$x$ drift}}& \multicolumn{2}{c|}{\textbf{$x$ diffusion}} &\multicolumn{2}{c|}{\textbf{$y$ drift}}& \multicolumn{2}{c}{\textbf{$y$ diffusion}} \\
\cline{2-9}
& Learnt & True& Learnt & True & Learnt & True& Learnt & True\\
\midrule  
1 & 0 & 0 & 1.0031 & 1 & 0 & 0 & 1.1847 & 1\\
$x$ & 0 & 0 & 0 & 0 & 25.0413 & 25 & 0 & 0\\
$y$ & -0.9972 & -1 & 0 & 0& 24.9859 & 25 & 0 & 0\\
$x^2$ & 0 & 0 & 0 & 0 & 0 & 0 & 0 & 0\\
$xy$ & 0 & 0 & 0 & 0 & 0 & 0 & 0 & 0\\
$y^2$ & 0 & 0 & 0 & 0 & 0 & 0 & 0 & 0\\
$x^3$ & 0 & 0 & 0 & 0 & 0 & 0 & 0 & 0\\
$x^2y$ & 0 & 0 & 0 & 0 & 0 & 0 & 0 & 0\\
$xy^2$ & 0 & 0 & 0 & 0 & 0 & 0 & 0 & 0\\
$y^3$ & 0 & 0 & 0 & 0 & -8.2930 & -8.3333 & 0 & 0\\
\bottomrule 
\end{tabular}
\end{table}

{\textbf{The invariant manifold}}

\begin{figure}[htp]
    \centering
    \includegraphics[width=8cm]{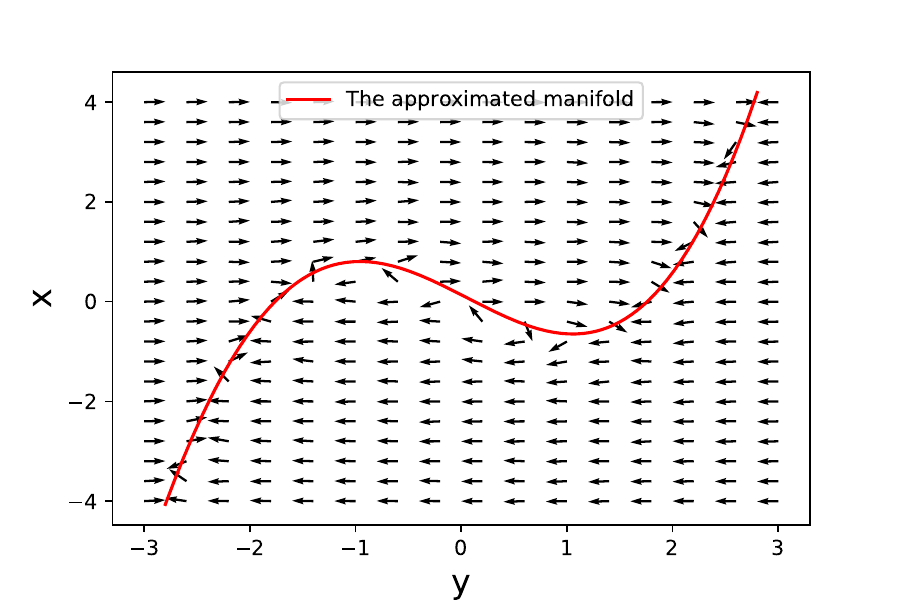}
    \caption{The approximated manifold(red curve) and the phase portrait(black arrow).}
    \label{SVDP mld}
\end{figure}

\begin{figure}[htp]
    \centering
    \includegraphics[width=8cm]{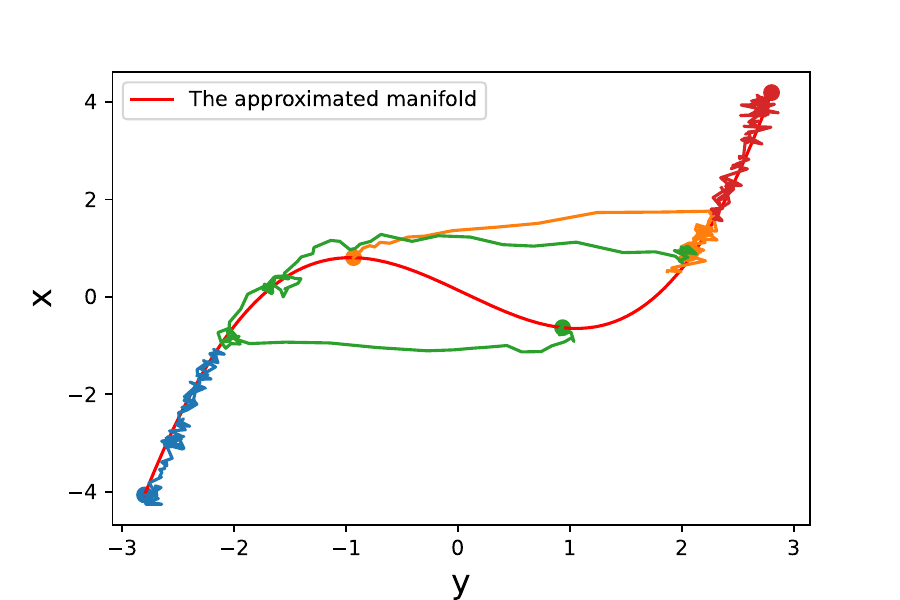}
    \caption{The approximated manifold (red curve) and four sample paths starting from the approximated manifold. The thick dots are starting points.}
    \label{SVDP traj}
\end{figure}

The snapshots of sample points at different time steps are similar to those of Example 1.  The mean absolute errors between two probability densities of adjacent snapshots are $0.024167$, $0.014271$, $0.012917$, $0.010278$ for $NT = 10$ and $20$, $NT = 20$ and $30$, $NT = 30$ and $40$, $NT = 40$ and $50$ separately.
The approximated manifold and the phase portrait of the stochastic Van der Pol system are plotted in {Fig.~$\ref{SVDP mld}$}. The phase portrait confirms that the points in the phase plane tend to the approximated manifold as time evolving. To better explain how well Auto-SDE approximate the invariant manifold, {Fig.~$\ref{SVDP traj}$} shows four sample paths starting from the approximated manifold selected randomly. It presents directly that the trajectories follow the directions in the phase portrait and are attracted to the approximated manifold. 
Moreover, the explicit expression of the approximated invariant manifold calculated by the snapshot data at $NT = 50$ can be written as $\hat{X}(y) = 0.0598 -0.9999y -0.0126y^2 + 0.3220y^3$.\\

\section{Conclusion and discussion}\label{CO}
In this paper, we design a general framework to learn effective reduced systems and dynamics, given short-term initial data and an unknown slow-fast stochastic dynamical system. To solve high dimensional data evolving over low dimensional space, we propose a neural network called Auto-SDE to approximate invariant manifold. The name of this network comes from considering our loss function evolves both autoencoder loss and the time evolutionary loss from a stochastic differential equation. First, we estimate the coefficients of the stochastic dynamical systems from the short time-interval pairwise data through the Kramers–Moyal formula and the polynomials. We then predict the dynamics over large time intervals and learn the invariant manifold as prediction time goes to long time period. Combining the slow equations and the invariant manifold, we can learn the effective reduced dynamics of the original systems. Three examples of nonlinear slow-fast systems are used to illustrate the validation of our method. For the first two example, when comparing some metrics of reduced system with the original slow variables, both the stochastic trajectories and the distributions show that our framework is promising. Furthermore, by varying the noise level, we demonstrate how noise affects the concentration of approximated dynamics near the manifold. For the third example, as the manifold cannot be explicitly expressed in terms of slow variables, we compare the manifold on top of phase portrait and show the various behaviors of sample paths near the invariant manifold. All in all, our framework based on Auto-SDE is general and accurate, and is effective to reduce high-dimensional systems to low dimensional space, as well as potentially contributing to the temporal nonlinear dimension reduction analysis of real-world data in science and engineering.

\section*{Acknowledgements}
We would like to thank Cheng Fang and Yubin Lu for helpful discussions. This work is supported by the Fundamental Research Funds for the Central Universities, HUST: 2022JYCXJJ058, National Key Research and Development Program of China 2021ZD0201300, National Natural Science Foundation of China 12141107, Fundamental Resaerch Funds for the Central Universities 5003011053 and Fundamental Research Funds for the Central Universities 5003011054.

\section*{Data availability}
The data that support the findings of this study are openly available in GitHub at 
https://github.com/LY-Feng/invariant-manifold.
\bibliographystyle{unsrt}
\bibliography{main}

\section{Appendix}
\noindent 

\begin{figure}[htp]
    \centering
    \includegraphics[width=8cm]{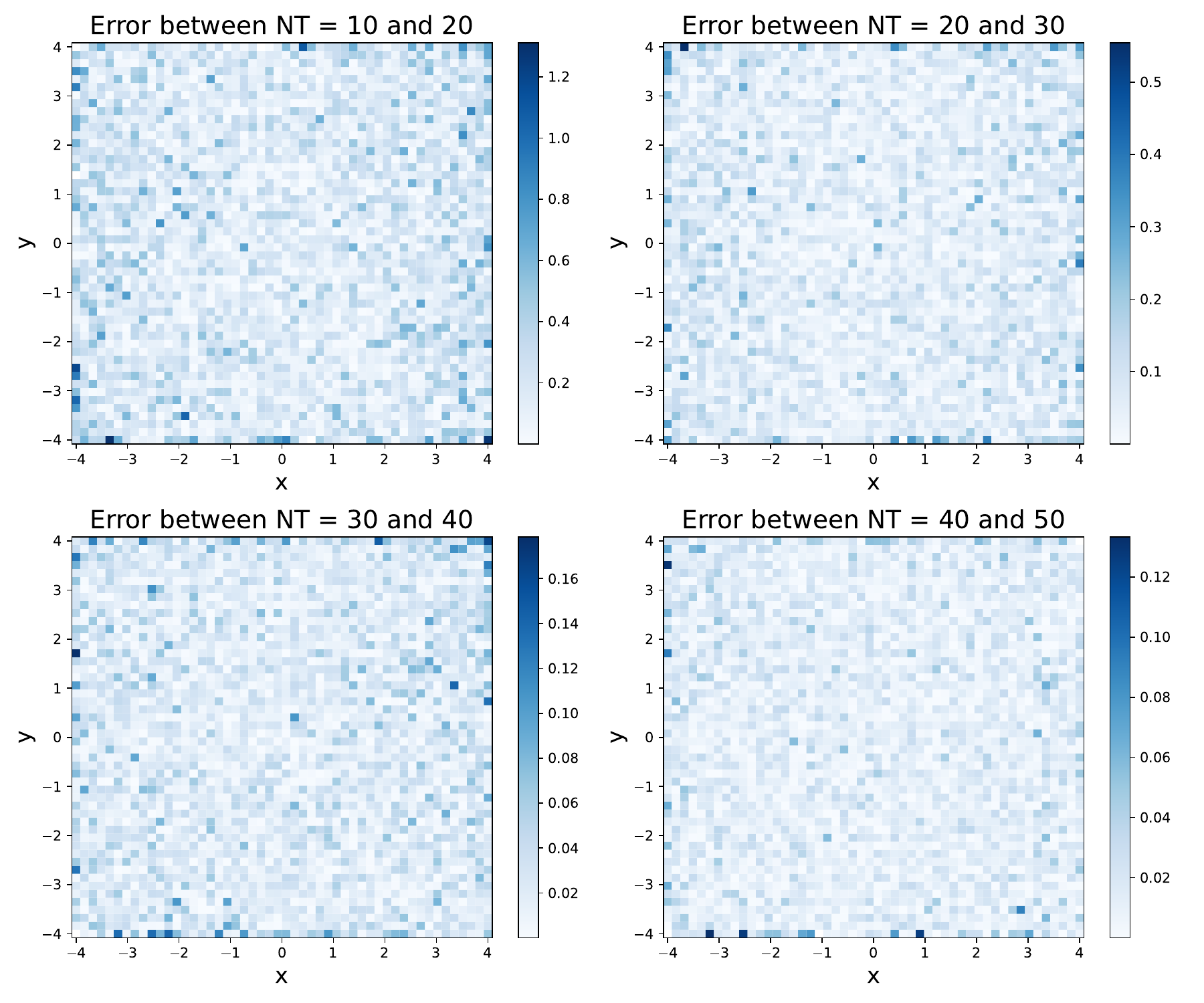}
    \caption{The absolute errors between two snapshots.}
    \label{Ex2_ABSERROR}
\end{figure}

The network keeps training and predicting recursively until the distribution of positions at some time step is similar to that of the former time step. To illustrate the convergence, we compute the absolute errors between two adjacent snapshots. In Example 2, the absolute errors of $y$ values on the plane $x_1Ox_2$ are shown in {Fig.~$\ref{Ex2_ABSERROR}$} and the mean values are $0.206738$, $0.070837$, $0.026578$, $0.015885$ for $NT = 10$ and $20$, $NT = 20$ and $30$, $NT = 30$ and $40$, $NT = 40$ and $50$ separately. For example, the the mean value between $NT = 10$ and $20$ is calculated by $\frac{\Sigma_{i=1}^{M}|y_{20}^i-{y}_{10}^i|}{M}$, where $M$ is the total number of $y$ values in the snapshot.

\end{document}